# Multilanguage Number Plate Detection using Convolutional Neural Networks

Jatin Gupta[1], Vandana Saini[2], Kamaldeep Garg[3]

*Chitkara University Institute of Engineering and Technology, Chitkara University, Punjab, India*

***Abstract —*** *Object Detection is a popular field of research for recent technologies. In recent years, profound learning performance attracts the researchers to use it in many applications. Number plate (NP) detection and classification is analyzed over decades however, it needs approaches which are more precise and state, language and design independent since cars are now moving from state to another easily. In this paperwe suggest a new strategy to detect NP and comprehend the nation, language and layout of NPs. YOLOv2 sensor with ResNet attribute extractor heart is proposed for NP detection and a brand new convolutional neural network architecture is suggested to classify NPs. The detector achieves average precision of 99.57% and country, language and layout classification precision of 99.33%. The results outperforms the majority of the previous works and can move the area forward toward international NP detection and recognition.*

***Keywords*** *- component; Number Plate Detection; Number Plate Classification; NPD; Yolo Detector; CNN; Deep Learning.*

## I. INTRODUCTION

Automatic number plate recognition (ANPR) is a set of techniques which use Number Plate Detection (NPD), character segmentation and character recognition on pictures to read automobile Number Plate (NP) numbers. It's also Known as Number Plate Detection and Recognition (NPDR). ANPR is employed in various real time applications such as railroad systems, electronic toll collection, traffic safety, management, etc., .

Lately, much research attentions have been attracted by object detection problem with the progress of learning [1]. Estimate places of objects found which class each item belongs to is called object detection and in a picture [2]. Object detection models' 3 phases are region selection, feature extraction and classification. The image is scanned by Area selection using a sliding window that is multi-scale as objects can appear with aspect ratios and assorted sizes [1].

Based on learning how the thing that is state-of-the-art detection algorithms deliver results on NP and NPD nation and design classifications. A distortion multi-orientation, lighting and multi-scale detection produces a struggle to NPD [15]. NPD using leaning is studied during the previous ten years. Writers in [16] suggested a CNN-based multi-directional (MD)-YOLO frame for

NPD however this system is hard to detect tiny NPs. In [17] R-CNN that was quicker was used by writers for the automobile region detection along with the NP in every automobile area was localized. The test is completed on the Caltech dataset using a precision of 98.39% and a recall of 96.83 percent. Plate pre-identification and A version YOLO-L is suggested in [18] where dimensions of NP candidate buys and the amount is chosen using + clustering algorithm using algorithm NPs from items and altered version. The largest NP named UFPR dataset was released in [19], they suggested character segmentation, vehicle detection, NPD, four phase NPDR system and character recognition. In NPD point, they utilized CR-CNN heart fast-YOLO and got a recall of 98.33 percent. Writers in [12] introduced a detailed and big Chinese NP dataset named dataset, they suggested an NPDR system. They suggested RPnet at NPD stage and they contrasted the detection average precision (AP) results with advanced detection techniques of SSD, YOLOv2 and Quicker R-CNN on 250k distinctive automobile NPs.

Emergence and discoveries of Convolutional Neural Networks (CNN) [9] attracted the investigators to utilize areas with (R-CNN) attributes for object detection problem [10]. CNN have a capability to learn attributes that are complicated with architectures and training algorithms let to find out object representations that are informative without needing to design attributes [11].

Generic object discovery is discovering objects from predefined courses with spatial location (bounding boxes) within a picture and it may be categorized into two kinds, area proposal and regression/classification based approaches [16]. Area proposal approaches consist of R-CNN [10], quick R-CNN [17], Quicker R-CNN [18] and Mask R-CNN [18].

Extractor of ResNet50. The dataset is called DeebNPD2020. After And classifier are analyzed on some datasets from that states and Arabia (KSA) to be discovered by one YOLOv2 detector with strong feature Our own dataset together. The suggested approach Segmentation of those detected NPs, the country will be recognized by a CNN, There are A few researches Completed on Multilanguage and Datasets used in also our dataset and this paper. Section 3 explains the In Tunisia the specimens are written in English although a mask R-CNN detector from Tunisia and USA for NPs of both Arabic and English figures along with also the country name is written in Arabic. Writers in [38] interested in English and Korean NPs and they utilized the expression multi-style to explain nation, speech and one or two lineup NP styles. Moreover, two line NP designs and one were needed to be categorized to rearrange the personalities in proper order in line with the country announcement.

NP and Vocabulary layouts with BR NPs. The proposed sensor Also it's thought of as the very first study concerning NPs from north and south Amerika, Europe and Middle East (TR, UAE and KSA) together. Brazil (BR), Europe (EU), Turkey (TR), United Arab Emirates (UAE) and Saudi Multinational NPs due to absence of international NP dataset however some research interested in creating a worldwide system [11]. The study [11] suggested an approach for multinational number plate detection for images with complex backgrounds in which the YUV colour space was utilized to found the rear vehicle lights the NP region was discovered by using a histogram-based strategy on the edge energy map. The dataset has NPs from America, China, Serbia, Italy, Pakistan, UAE and Hungary that the dataset had single-line NPs and a accuracy of 90% was assessed. Writers in used VGG to classify the enrollment state of NPs from Lithuania, Latvia, Estonia, Russia, Sweden, Poland, Germany, Finland and Belarus. Recent research used tiny YOLOv3 to discover NPs from South Korea, Taiwan, Greece, USA, and Croatia [13]. But some recent researches interested in multinational NPs but they analyzed their sensors rather than collecting them in 1 test set .

Closes the difference of NPs, Multilanguage and multinational layouts in 1 system Lately, Multilanguage NPs bring some intention. Writers in [7] Approach in details. Section 4 presents a series of experimental outcomes and discussions. Finally, Section 5 concludes that the main factors of the whole In this paper we worried NPs from USA, work

## II. DATASET

### A. NP Datasets available in the Litrature

In this research we include some NP datasets available TR, UAE and KSA there are Online in pervious researches from our concerned countries for comparison. For benchmarking. TABLE I. summarizes the arrangement of these datasets.

TABLE I.     DATASETS

| Dataset | Year | # of images | Accuracy % | Country |
|---|---|---|---|---|
| **Error! Reference source not found.** | 2011 | 128 | --- | India |
| [10] | 2012 | 861 | 76.2 | Nepal |
| [12] | 2015 | 906 | 79.5 | USA |
| [23] | 2017 | 971 | 79 | Africa |
| [24] | 2018 | 751 | --- | Greece |
| [28] | 2017 | 78100 | 86.5 | Czech |
| [27] | 2018 | 4789 | 88.33 | India |

### B. Our NP dataset

This research introduces a dataset which is collected from our personal camera from India and a few sites. There's no available Hindi NPs datasets therefore we gathered our NP pictures from internet for academic study. TABLE II. Shows a summary about our images and annotated NP images.

Some NPs from different countries with one-line and two-line layouts from our dataset called DeebNP2020 are shown in Fig. 1.

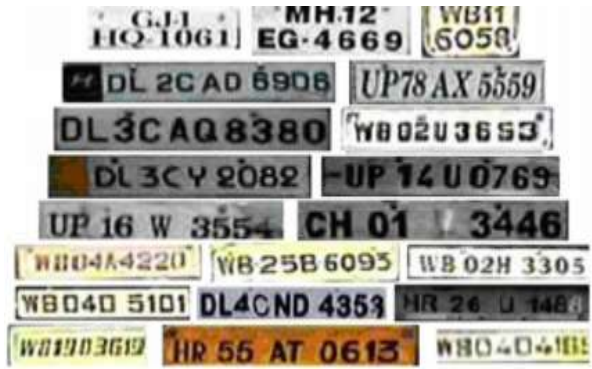

Fig. 1.     Example of some NPs and layouts.

NP classification dataset has 11 classes Sensor. ResNet50 structure is explained in TABLE III. Is to detect a NP in a picture and the second problem is to harvest the detected NP and categorize its own nation, language and layout. In the literature there's some researchers solved both of problems simultaneously using one detector but for either just nation, language or layout.

Our approach is designed using YOLOv2 detector which ResNet50 A. Number This study concerns two problems; the Issue Network is utilized as the center CNN for its shown in Fig. 1. and TABLE III. Describes the construction of the NP dataset. The entire number of our NP classification dataset is 35000 NP images in which they are gathered in the countries from three regions Europe, America and Middle East.

TABLE II.   RESNET50 ARCHITECTURE

| Layer | Output size | Filters & size |
|---|---|---|
| Input | $228 \times 228 \times 3$ | -------- |
| Conv1 | $116 \times 116 \times 32$ | $124 \times 124 \times 3$, stride 2 |
| Max pooling | $128 \times 128 \times 64$ | $3 \times 3$ max pool, stride 2 |
| Conv2 | $64 \times 64 \times 256$ | $\begin{bmatrix} 1 \times 2, 64 \\ 3 \times 3, 64 \\ 1 \times 2, 256 \end{bmatrix} \times 3$ |
| Conv3 | $128 \times 128 \times 512$ | $\begin{bmatrix} 1 \times 1, 128 \\ 3 \times 3, 128 \\ 1 \times 1, 512 \end{bmatrix} \times 6$ |
| Conv4 | $8 \times 8 \times 1024$ | $\begin{bmatrix} 1 \times 1, 256 \\ 3 \times 3, 256 \\ 1 \times 1, 1024 \end{bmatrix} \times 8$ |
| Conv5 | $16 \times 16 \times 1048$ | $\begin{bmatrix} 8 \times 1, 512 \\ 3 \times 3, 512 \\ 4 \times 1, 2048 \end{bmatrix} \times 6$ |
| Average pooling | $10 \times 10 \times 1048$ | $7 \times 7$ |
| Fully Connected | $12 \times 12 \times 1200$ | 2000 |
| Softmax | $4 \times 4 \times 800$ | -------- |

INSPIRE-AVR and MESSIDOR database described, are revealed and compared with other research. Elaborates the materials and methods [9] presented a vessel segmentation Method based on a multi-scale linear structure enhancement and the second order local entropy thresholding. The procedure is tested on 74 images with 80% accuracy. The restriction of their job was that the AVR calculation has been done only on area (0.5-1) optic disc diameter (DD) which may give improper results. [10] Introduced a hessian-based vessel segmentation method along with thresholding and achieved an accuracy of 93.71% and 93.18percent for DRIVE and STARE respectively. In AVR calculation, area of interest (ROI) is highly determined by optic disc detection but they didn't supply a very clear idea of sensory disk detection procedure which was used in their work. [11] Introduced a new process of AVR computation. They tested their method on 58 fundus images of VICAVR and 40 images of DRIVE database and attained a precision of 96.5% and 98 percent respectively. The limitation of their work was that they not present sufficient description of the results. [12] Presented a vessel tracking algorithm. The process was tested on 50 images. Some of the boats were overlooked by the tracing procedure therefore gave AVR values that were unsatisfactory. [13] Utilized a blood vessel detection procedure using median filter and top hat transform on 76 retinal fundus pictures of VICAVR database along with 25 images which were clinically acquired. In spite using a large dataset information is provided concerning variety of AVR values. Thus, these methods can't be used to see the retinopathy. In the present work, there is a new approach designed to extract retinal vasculature structure using hat that was dual transform. The simplicity of the method and precision that is higher on relatively large database with number of fundus images demonstrates that the work may overcome the limitation of their prior functions.

Used in Section III the experimental results on; for automatic detection of HR Calculations; Section IV outlines the findings along with the pointers to prospective The paper is organized as follows: Section II.

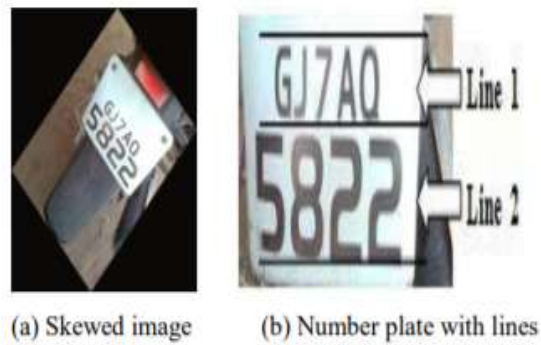

Fig. 2.   Number Plate System.

YOLOv2 detector divides the input image into $S \times S$ grid where $S$ is the output feature map size of YOLOv2 core Resnet40 which is $42 \times 42$ in our approach (the output of Conv4 layer) then anchor boxes is resized down by forward propagation size. YOLOv2 uses $A$ anchor boxes to predict bounding boxes with four coordinates, confidence and C class probabilities [20], the number of filters is given by

$$filters = (C + 5) \times A \qquad (1)$$

We analyzed our NP sizes in DeebNPD2020 dataset to select our anchor boxes. We used pyramid of anchors method in Faster R-CNN [16] to analyze and select our anchor boxes.

As shown in Fig. 3. NP sizes span from 10 to 670 so to select anchor boxes of high intersection of union (IOU) we used 6 minimum NP sizes defined as $hieght \times width$ of $10 \times 10$ , $10 \times 20$ , $10 \times 30$ , $10 \times 40$, $10 \times 50$ and $30 \times 14$ with pyramid level of 15 and anchor box pyramid scale of 1.3 to obtain 90 anchor boxes where minimum IOU of 0.625 and mean IOU of 0.85 are obtained. According to (1) our last YOLOv2 layer has $(1 + 5) \times 90 = 540$ filters as shown in Fig. 2.

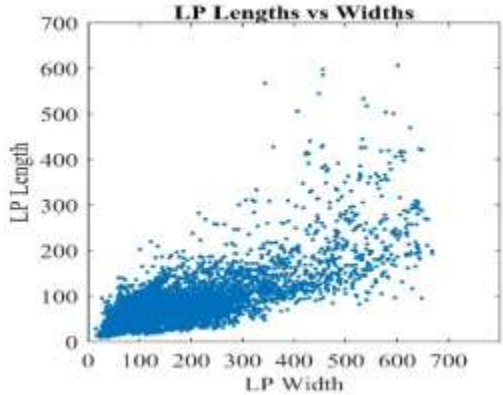

Fig. 3.    NP sizes in DeebNPD2020 dataset.

### C. Number Plate Classification

As our research concerns NP multinational, multilanguage and layout classification, we designed a simple CNN and we compare its accuracy to VGG [10].

TABLE II.    PROPOSED CNN DESIGN

| Layer | Filters & size | Output | Learnable parameters |
|---|---|---|---|
| Input | -------- | 224×224×3 | ------------- |
| Conv1 | 5×5×32 | 220×220×32 | 2496 |
| Conv2 | 5×5×32 | 216×216×32 | 25696 |
| Maxpooling | 2×2 | 108×108×32 | ------------- |
| Conv3 | 5×5×64 | 104×104×64 | 51392 |
| Conv4 | 5×5×64 | 100×100×64 | 102592 |
| Maxpooling | 2×2 | 50×50×64 | ------------- |
| Conv5 | 5×5×96 | 46×46×96 | 153888 |
| Conv6 | 5×5×96 | 42×42×96 | 230688 |
| Maxpooling | 2×2 | 21×21×96 | ------------- |
| Conv7 | 5×5×128 | 17×17×128 | 307584 |
| Conv8 | 5×5×128 | 13×13×128 | 409984 |
| Maxpooling | 2×2 | 6×6×128 | ------------- |
| Conv9 | 5×5×256 | 2×2×256 | 819968 |
| Conv10 | 2×2×512 | 1×1×512 | 525824 |
| Fully Connected | 9 | 1×1×9 | 4617 |
| Softmax | -------- | 1×1×9 | ------------- |

The proposed design has a total number of learnable parameters of 2634729 (2.635M) that is less than VGG learnable parameters of 138 million parameters. Every convolutional layer followed by batch normalization (BN) layer and ReLU **Error! Reference source not found.** non-linear activation layer, BN normalizes the input batch mean and standard deviation then scales and shifts it by learnable scale and shift parameters **Error! Reference source not found.**. As shown in TABLE II. all the convolution kernels have a size of 5 × 5 with stride 1 and no padding; this means every convolutional layer makes a dimension shrinkage of 4 pixels. The dimension of the output feature map can be computed according to (2).

$$W_{out} = \frac{W_{in} - W_k + 2P}{W_s} + 1 \qquad (2)$$

Where $W_{out}$ is the output feature map width, $W_{in}$ is the input feature map width, $W_k$ is the kernel width, $P$ is the padding, and $W_s$ is the kernel stride in the horizontal direction. For input/output height dimension (2) can be applied too but with $H$ instead of $W$.

In the training process, we used stochastic gradient decent with momentum (SGDM) **Error! Reference source not found.** with 10 epochs, initial learning rate (LR) drop factor by half every 2 epochs and shuffle the training set every epoch. In YOLOv2 training the mini-batch size is only 6 images due to memory constraints and LR is $1 \times 10^{-5}$ while NP classification CNN mini-batch size is 120 images and LR is $2.5 \times 10^{-2}$. After getting the first results a model parameters tuning is applied by continue training but on ADAM **Error! Reference source not found.** with larger mini-batch size and very small learning rate started by $1 \times 10^{-5}$, then multiply the batch size by 2 and LR by 1/2 every 10 epochs as long as the test error has improvement.

### III. RESULTS & DISCUSSION

This research used Matlab2019B environment with GeForce 1060 6G GPU with compute capability 6.1 for both training and testing processes. The evaluation criteria for both NPD and NP classification will be described in next sections.

### A. NPD

The NP detection evaluation is done using precision (P), recall (R) and average precision (AP) on detected NP bounding box with overlap greater than IOU=0.5 between the predicted and the ground truth bounding box. Precision is the percentage of the number of correctly detected NPs over the total number of detected NPs while the recall is the percentage of the number of correctly detected NPs over the total number of ground truth NPs and AP is the area under the precision recall curve. Precision and recall are defined as follows:

$$Precsion = \frac{TP}{TP + FP} \qquad (3)$$

$$Recall = \frac{TP}{TP + FN} \qquad (4)$$

where TP is true positive, FP is false positive and FN is false negative.

TABLE III. Shows the proposed detector architecture AP performance compared to previous works performances described in the literature. The proposed detector outperforms the previous researches in terms of AP performance. However, in **Error! Reference source not found.** they consider only character recognition accuracy and NPD processing time but they do not mention the NPD stage performance while in **Error! Reference source not found.** they just evaluate the accuracy for the detected NPs over all NPs in a private dataset but in **Error! Reference source not found.** they just evaluate NPD precision. The proposed detector processing time is lower than **Error! Reference source not found.** because they used just 189 NPs in their dataset for USA and Korea.

TABLE III. PROPOSED APPROACH NPD RESULTS.

| Approach | Detector | AP | Processing Time in seconds |
|---|---|---|---|
| **Error! Reference source not found.** | image processing | 90.4% accuracy | 0.25 |
| **Error! Reference source not found.** | VGG+ LSTM | 98.07% | Not reported |
| **Error! Reference source not found.** | image processing + Alexnet +SVM | 99.03% precision | 0.16 |
| **Error! Reference source not found.** | KNN | ---- | 0.027 |
| Proposed | ResNet40+ Yolov2 | **99.57%** | 0.09 |

Some researchers trained and tested detectors on every dataset separately but our proposed detector has common learnable parameters trained on all datasets together and tested on everyone separately. **Error! Reference source not found.**shows the performances in terms of detection precision, recall and average precision.

To make a fair comparison in **Error! Reference source not found.**it is needed to train our detector on every dataset separately. However, our detector has the best recall rate and AP over all datasets due to our large number of different NPs used in our dataset. In those approaches, large number of East Asian NPs and small number of English NPs where used with background color differences.

One and two line NP layouts classification is studied in **Error! Reference source not found.** with classification result combined in character recognition stage for multinational Korean, Taiwanese, Chinese and English NPs characters.

The proposed approach is also tested over the interested countries and TABLE IV. Shows the AP results per country.

TABLE IV. AP RESULTS ON DeebNPD2020 DATASET PER COUNTRY

| Dataset | US | EU | IND | ARAB | SA |
|---|---|---|---|---|---|
| Proposed | 98.48% | 99.21% | 99.95% | 99.43% | 98.17% |

### B. NP Calssification

The proposed CNN for classifying issuing country, language and layout of NPs is tested in terms of accuracy. There is 11 NP classes from 6 countries as shown in Fig. 1. and described in **Error! Reference source not found.**where the language inferred from the country class.

TABLE V. PROPOSED CNN NP CLASSIFICATION ACCURACY

| CNN architecture | Accuracy |
|---|---|
| IC104T | 92.71% |
| Proposed CNN Method | 99.12% |

## IV. CONCLUSION

This research concerned the NPD and classification of NPs of both Multilanguage, multinational and layout from India, USA, EU, TR, KSA and UAE. This study considered to suggest a sensor that can detect NP irrespective of language its nation or layout with another stage to recognize that the issuing nation, language and one or 2 line NP design. Discovering terminology and the nation is required to build an worldwide ANPR system which it is not exist while design classification is needed to browse the detected personalities in correct order. The classification and detection results with profound learning are very promising. A new gate is opened to design a international ANPR system. In addition NP datasets called close the gap in the area and DeebNPD2020 and DeebNP2020 are introduced to make new tests. In future researches can be conducted to improve the operation or test a brand new approaches like evaluation the same sensor with NP classes in one point, change the input size to meet standard frame sizes or decrease CNN scales to make it. On the flip side, an end-to-end training procedure can be proposed to test the entire system as a unified ANPR version. Model.